\documentclass[letterpaper, 10 pt, conference]{ieeeconf} 
\IEEEoverridecommandlockouts

\usepackage{epsfig}
\usepackage{algorithm}
\usepackage{algorithmic}
\usepackage{amsmath,graphicx}            
\usepackage{amsfonts}
\usepackage{color}
\usepackage{xfrac}
\usepackage{caption}
\usepackage{subfigure}
\usepackage{amssymb}
\usepackage{bm}

\newcommand{\Tcal}{\mathcal{T}}

\newcommand{\Scal}{\mathcal{S}}
\newcommand{\Ocal}{\mathcal{O}}
\newcommand{\Acal}{\mathcal{A}}

\newcommand{\obs}{\boldsymbol{o}}

\newcommand{\pibs}{\boldsymbol{\pi}}

\title{\LARGE \bf
MAPPER: Multi-Agent Path Planning with Evolutionary Reinforcement Learning in Mixed Dynamic Environments}

\author{Zuxin Liu$^{1}$, Baiming Chen$^{2}$, Hongyi Zhou$^{1}$, Guru Koushik$^{1}$
, Martial Hebert$^{3}$ 
and Ding Zhao$^{1,*}$
\thanks{* Corresponding author: Ding Zhao. Email: ({\tt\small dingzhao@cmu.edu})}%
\thanks{$^{1}$Department of Mechanical Engineering, Carnegie Mellon University, USA.}%
\thanks{$^{2}$School of Vehicle and Mobility, Tsinghua University, China.}%
\thanks{$^{3}$The Robotics Institute, Carnegie Mellon University, USA.}%
}
\begin{document}
\maketitle
\thispagestyle{empty}
\pagestyle{empty}
\begin{abstract}
Multi-agent navigation in dynamic environments is of great industrial value when deploying a large scale fleet of robot to real-world applications. This paper proposes a decentralized partially observable multi-agent path planning with evolutionary reinforcement learning (MAPPER) method to learn an effective local planning policy in mixed dynamic environments. Reinforcement learning-based methods usually suffer performance degradation on long-horizon tasks with goal-conditioned sparse rewards, so we decompose the long-range navigation task into many easier sub-tasks under the guidance of a global planner, which increases agents' performance in large environments. Moreover, most existing multi-agent planning approaches assume either perfect information of the surrounding environment or homogeneity of nearby dynamic agents, which may not hold in practice. Our approach models dynamic obstacles' behavior with an image-based representation and trains a policy in mixed dynamic environments without homogeneity assumption. To ensure multi-agent training stability and performance, we propose an evolutionary training approach that can be easily scaled to large and complex environments. Experiments show that MAPPER is able to achieve higher success rates and more stable performance when exposed to a large number of non-cooperative dynamic obstacles compared with traditional reaction-based planner LRA* and the state-of-the-art learning-based method.
\end{abstract}

\section{INTRODUCTION}
\label{sec:intro}
Driven by the need for flexible and efficient manufacturing, an increasing number of affordable mobile robots are expected to be deployed in warehouse environments for transportation purposes. One key component to support the applications of large scale robots is the multi-agent path planning technology. Many research efforts have been devoted to this field in recent years from different perspectives. 

Generally, multi-agent planning methods can be classified into two categories: centralized methods and decentralized methods. When all the moving agents' intentions (e.g. future trajectories, goals) are known in a static environment, a centralized planner could generate collision-free paths for all the agents \cite{sharon2015conflict}. However, the computational burden may be a significant concern as the number of agents grows, and the agent's performance may degrade when exposed to unknown dynamic objects  \cite{mellinger2012mixed}. Besides, in practice, centralized methods heavily rely on stable and fast communication networks and powerful servers, which would be costly to be deployed in large scale environments with a large number of robots. Therefore, in this paper, we focus on decentralized methods, where reliable communication can not be established between agents.

For decentralized methods, each agent independently makes decisions based on its own local observations and policies. A natural question is: what should the agent know and assume about other agents or dynamic obstacles around it? Some approaches assume all obstacles are static and re-plan at a high frequency to avoid collision \cite{koenig2005fast}, while other people assume homogeneous policy for agents and constant velocity for dynamic obstacles \cite{van2008reciprocal}. However, we argue that in practice, it is difficult to perfectly estimate neighbouring decision-making agents' intentions without communication. Therefore, instead of using traditional path planning procedures, some recent approaches use reinforcement learning to solve robot the navigation problem by implicitly learning to deal with such interaction ambiguity with surrounding moving obstacles \cite{chen2017socially, everett2018motion, mehta2016autonomous, sartoretti2019primal}.

Though learning-based approaches have shown great potential to model complex interactions in dynamic environments, most of them make assumptions about the homogeneity or motion models of surrounding moving obstacles \cite{long2018towards, chen2017decentralized}. In this paper, we focus on planning in mixed dynamic environments where moving obstacles can either be cooperative or non-cooperative. Also, inspired by state-of-the-art trajectory prediction methods \cite{cui2019multimodal}, we propose an image-based spatial-temporal dynamic obstacle representation, which doesn't need explicit motion estimation and can be generalized to the arbitrary number of agents.

Reinforcement learning agent is usually difficult to achieve satisfying performance in long-horizon tasks with sparse rewards, as in the long-range goal-conditioned navigation problem \cite{eysenbach2019search}.
Therefore, one insight in this paper is using mature planning methods to guide the reinforcement learning-based local planning policy. In this way, agents can learn from complicated local interaction with dynamic obstacles while persistently moving towards a long-range goal. In addition, to ensure the multi-agent training stability and performance, we propose an evolutionary reinforcement learning method that can be easily scaled to large and complex environments. 

The major contributions of this paper are:
\begin{enumerate}
    \item We investigate the multi-agent path planning problem in mixed dynamic environments without the homogeneity assumption. 
    To model the dynamic obstacles' behavior, we propose an image-based representation which improves our agents' robustness to handle different types of dynamic obstacles. 
    \item We decompose a difficult long-range planning problem into multiple easier waypoint-conditioned planning tasks with the help of mature global planners. Experiments show that this approach can greatly improve the performance of reinforcement learning-based methods.
    \item We propose an evolutionary multi-agent reinforcement learning approach that gradually eliminate low-performance policies during training to increase training efficiency and performance.
\end{enumerate}

The structure of this paper is as follows. Section~\ref{sec:related} introduces related works about multi-agent path planning in dynamic environments. Section~\ref{sec:background} provides some preliminaries for our problem formulation. The detail of our multi-agent path planning with the evolutionary reinforcement learning (MAPPER) approach is presented in section~\ref{sec:approach}, followed by section~\ref{sec:experiment} which shows the experiment results of MAPPER in various grid world simulations with mixed dynamic obstacles. Finally, we give a brief conclusion in section~\ref{sec:conclusion}.
\section{Related Work}
\label{sec:related}

\subsection{Decentralized Path Planning in Dynamic Environment}
Decentralized planning methods can be broadly classified into two categories: reaction-based and trajectory-based.
For reaction-based approaches, we need to specify avoidance rules for dynamic obstacles and re-plan at each step based on the rules, such as D* lite \cite{koenig2005fast} and velocity obstacle (VO) based methods \cite{snape2011hybrid, bareiss2015generalized}. 
Trajectory-based approaches usually estimate surrounding dynamic objects' intentions and then search collision-free paths in the space-temporal domain \cite{fan2018baidu}. These methods either require perfect information of surrounding dynamic obstacles (e.g. velocities and positions) or assume that all the moving agents adopt the same planning and control policy so that they are homogeneous \cite{van2011reciprocal, van2008reciprocal}. However, such assumptions may not hold in many real-world applications when involved with sensing uncertainty and heterogeneous moving obstacles, such as pedestrians. Beside, increasing local interaction complexity may lead to oscillation behaviors or \textit{freezing robot problem} \cite{trautman2010unfreezing}. Also, in practice, VO-based and trajectory-based approaches usually have several components to process sensor data, such as object-intention estimation and trajectory prediction. Each component may have many hyper-parameters that are sensitive to environment settings, which need extra human efforts to tune.
In order to reduce the amount of hand-tuning parameters and deal with sensing uncertainties, some researchers proposed learning-based methods to solve the planning problem.

\subsection{Reinforcement Learning-based Planning}
Reinforcement learning-based collision-avoidance algorithms for the single-agent case have been extensively studied in recent years. Deep neural networks are usually used to approximate agent's policy and value function. Some people propose to learn the navigation policy in a completely end-to-end fashion, which directly maps raw sensor data to the agent's action \cite{pfeiffer2017perception,tai2017virtual}. However, we believe that extracting object-level representation can improve the policy generalization ability, because different sensor data sources may encode the same object-level information. Chen et al. first estimate dynamic obstacle's physical states (e.g. velocity and positions) under certain motion model assumptions, and then feed them into neural network to obtain future actions~\cite{chen2017socially, chen2017decentralized} . However, the agents' number is restricted and can hardly be deployed in large scale environments. \cite{everett2018motion} addresses the problem of a varying number of dynamic obstacles with LSTM and removes the homogeneity assumption for surrounding agents. However, it still requires explicit estimation of surrounding agents' states and suffers performance degradation in tasks with a large number of agents. For multi-agent case, PRIMAL \cite{sartoretti2019primal} is the most similar work with ours, which also uses image-based representation and target goal as input sources. However, non-cooperative dynamic obstacles and temporal information are not considered in their work. Besides, their centralized training approach takes a long time even with the help of imitation learning. Our approach differs from theirs in that: 1) we encode both spatial and temporal information of surrounding obstacles in observation representations; 2) we consider planning in mixed dynamic environments; 3) we propose a decentralized evolutionary training method which can converge much faster and can be generalized to arbitrary number of training agents; 4) we use mature global planner to guide the local policy to solve long-range navigation problem. We will use the reinforcement learning method in PRIMAL as an experiment baseline in section~\ref{sec:experiment}.

\section{Background}
\label{sec:background}

\subsection{Problem Formulation}
We model the multi-agent planning problem under the Markov decision processes (MDPs) framework. An $N$-agent MDPs can be represented by the state space $\Scal$, which describes all the possible state configurations of the agents, the action space $\Acal_1, ..., \Acal_N$, which defines each agent's possible actions, and the observation space for each agent $\Ocal_1, ..., \Ocal_N$. In this paper, we consider the partially observable situation, which means each agent can not observe all the state. The agent $i$ receives its own observation $\obs_i:\Scal \mapsto \Ocal_i$ and produces an action from its stochastic policy $\pibs_{\theta_i}:\Ocal_i \times \Acal_i \mapsto [0,1]$, where the policy is parameterized by $\theta_i$. All the agents' actions will produce new state that follows the state transition function $\Tcal : \Scal \times \Acal_1 \times ... \times \Acal_N \mapsto \Scal$. For each time step, agent $i$ will receive rewards based on state and its action $r_i:\Scal \times \Acal \mapsto \mathbb{R}$. The initial states is determined by the distribution $\rho:\Scal \mapsto [0,1]$. The objective is to maximize the expected return $R_i = \sum_{t=0}^{T} \gamma^tr_i^t$ of each agent $i$, where $\gamma$ represents the discount factor and $T$ is the time length. The detail representations of observation space, action space and rewards will be introduced in section~\ref{sec:approach}.

\subsection{Advantage Actor Critic Algorithm}
We use advantage actor-critic (A2C) method \cite{mnih2016asynchronous} as the basis of our multi-agent evolutionary reinforcement learning framework to solve the planning problem. A2C uses stochastic policy, which is essential in our multi-agent scenario because the equilibrium policies in multi-agent MDPs are usually stochastic \cite{bucsoniu2010multi}. Additionally, policy gradient-based methods usually have better convergence property than value-based methods \cite{mnih2016asynchronous}. The objective of A2C is to find the policy $\pibs_{\theta}( a |\obs)$ that can maximize the expected return $\mathbb{E}_{\pi_\theta}R(\tau) = \mathbb{E}_{\pi_\theta}\sum_{t=0}^{T} \gamma^tr(\obs_t, a_t)$ over the episode $\tau = (\obs_0, a_0, \dots, \obs_T, a_T)$, where $a$ is the action and $\obs$ is the observation. Given this objective function, the policy gradient can be computed as:
\begin{equation}
    \nabla_\theta J(\theta) = \mathbb{E}_{\pibs_\theta}[\nabla_\theta \log \pibs_{\theta}( a |\obs) R(\tau)]
\end{equation}
To reduce the gradient variance, we employ a value function $V^{\pibs_{\theta}}(\obs)$ as the baseline and replace the expected return $R(\tau)$ with an advantage function $A^{\pibs_{\theta}}(\obs, a)$. Then we can rewrite the gradients as:
\begin{equation}
    \nabla_\theta J(\theta) = \mathbb{E}_{\pibs_\theta}[\nabla_\theta \log \pibs_{\theta}( a |\obs) A^{\pibs_{\theta}}(\obs, a)],
\end{equation}
where the advantage function $A^{\pibs_{\theta}}(\obs, a)$ has an unbiased estimation:
\begin{equation}
    A^{\pibs_{\theta}}(\obs, a) = \mathbb{E}_{\pibs_\theta}(r + \gamma V^{\pibs_{\theta}}(\obs^\prime) | \obs, a) - V^{\pibs_{\theta}}(\obs)
\end{equation}
The policy $\pibs_\theta$ is usually termed as the \textit{actor} to produce actions based on current observations, and the value function $v^{\pibs_{\theta}}$ is the \textit{critic}, which is used to estimate the advantage function $A^{\pibs_{\theta}}$ that indicates the quality of produced actions. In this paper, we approximate the policy and value function with neural networks, which will be introduced in section~\ref{archi}.

\section{Approach}
\label{sec:approach}

This section shows how the multi-agent path planning problem is modeled into an evolutionary reinforcement learning framework. We firstly introduce the observation representation, action space, and reward design of each agent. Then, we detail the model architecture and training procedures.

\subsection{Observation Representation}
\label{obs}
In many real-world mobile robot applications, people usually use the single beam LiDAR for localization and obstacle detection purposes, which is cheap and reliable \cite{grisetti2007improved, pierson2019dynamic}. A common map representation based on the LiDAR data is called the cost map, which discretizes a 2D map into a fixed resolution grids and assigns a cost to each grid \cite{reid2013cooperative}. The cost and obstacle information can be continuously updated by the local observation of sensor data. Therefore, to mimic such common map representations in practice, we consider a partially observable grid world environment, where each agent has its own visibility that limited by the sensing range and there is no communication between agents. We argue that such a fully decentralized partially observable setting is feasible and important if we need to deploy our approach to the real-world with large scale robots. We assume that each agent is able to detect and distinguish surrounding agents and dynamic objects within its sensing range and estimate their relative positions. Also, we assume that each agent can access the static environment map so that it can plan a trajectory in this map.

\begin{figure}[htb]
\centering
\includegraphics[width=8.3cm]{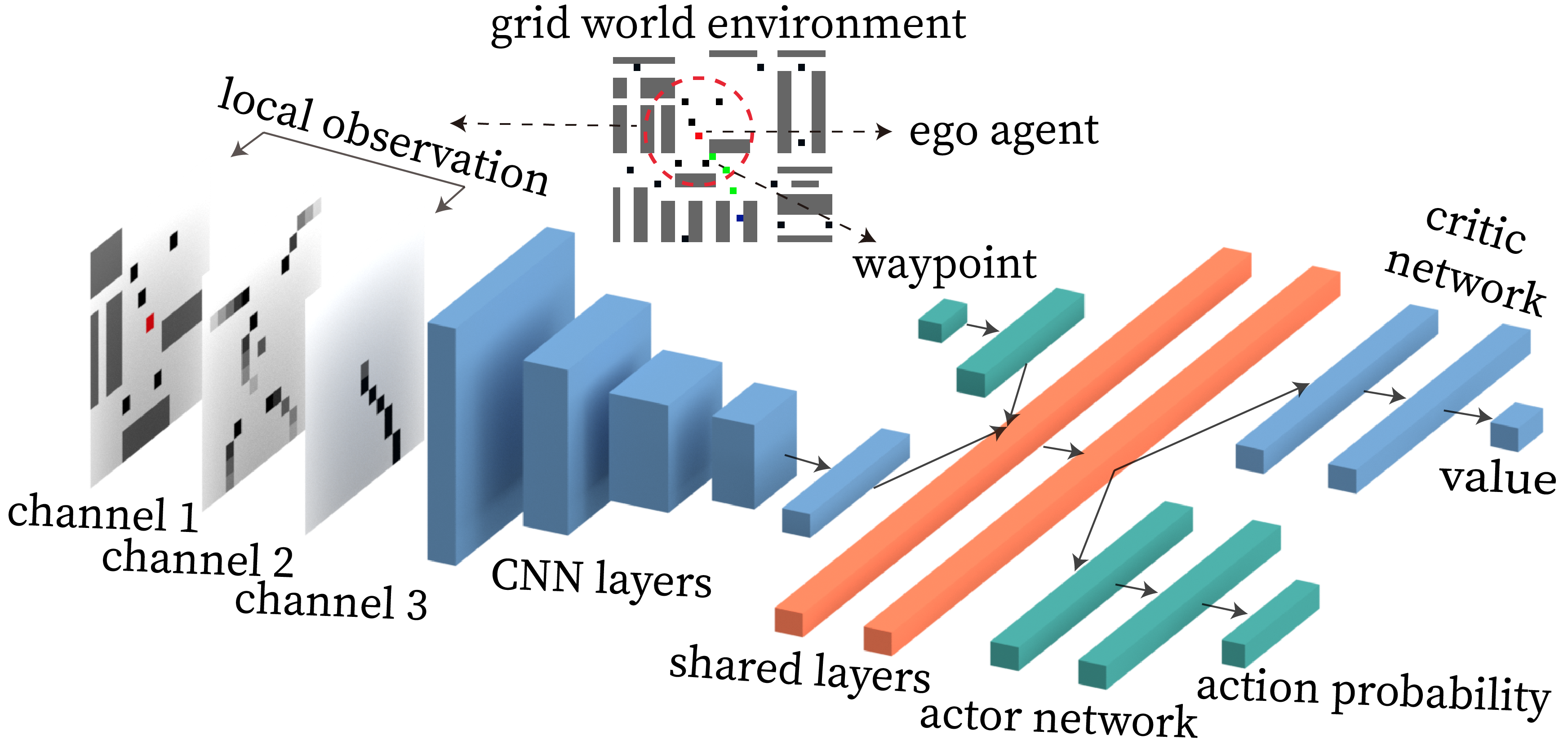}
\caption{MAPPER model architecture overview.}
\label{fig:framework}
\end{figure}

We split the observations into three channels to encode different types of information. As shown in Fig.~\ref{fig:framework}, the first channel stores current observed static obstacles, surrounding agents and dynamic objects' positions, which are represented by different values. This channel is the basic reflection of sensing data and is corresponding to the cost map representation, which could be used in many traditional search-based planning algorithms \cite{koenig2005fast}. The second channel is the trajectory of surrounding agents and dynamic obstacles, which encodes the time sequence information. Inspired by the state-of-the-art trajectory prediction method in the autonomous vehicle field, we encode the trajectory with different grayscales in time \cite{cui2019multimodal}. For example, the point on a trajectory in the earlier time has a smaller value than the later one. The third channel is the reference path planned by a global planner based on the static environment map. The reference path update frequency could be much lower than our reinforcement learning-based local planner. We will demonstrate the importance of those observation representations in the experiment section~\ref{sec:experiment}.

\subsection{Action Space}
In this paper, we consider an 8-connected grid environment, which means the agent can move to 8 directions: south, north, west, east, southwest, northwest, southeast and northeast. The agent can also choose to wait at the current grid. Thus, the action space contains 9 discrete options in total. At each time step, the agent will move one step following the direction that is selected. However, if the target grid is already occupied, the agent will not be able to move and will stay at the current position.

\subsection{Reward Design}
The objective of robot navigation is to reach the goal position with minimum steps while avoiding collision with obstacles. Therefore, the first part of the reward consists of step penalty $r_s$, collision penalty $r_c$ and goal-reaching reward $r_g$. 
To encourage exploration, we penalize slightly more for waiting than moving if the agent has not reached the goal. A similar training trick is also used in \cite{sartoretti2019primal}. To prevent agents from adopting oscillating policies, we set penalty $r_o$ to agents that return to the positions they come from last time. The detailed values of these reward components in our experiment can be found in Table~\ref{reward}. 

Since our local planning policy is guided by a reference path planned by global planner, we introduce an additional off-route penalty $r_{f}$ if the agent deviates from the reference path. The intuition is that if there are no dynamic obstacles around the agent, it should be able to follow the reference path. To obtain the off-route penalty,
we need to calculate the Euclidean distance between the agent's position and the closest point's position on the reference path. Denote the position of the agent as $\boldsymbol{p}_a \in \mathbb{R}^2$. Denote the reference path as a set of coordinates $\mathcal{S} = \{\boldsymbol{p}_{start},...,\boldsymbol{p}_{goal}\}$, 
the penalty is calculated by
$
    r_{f} = -\min_{\boldsymbol{p} \in \mathcal{S}} ||\boldsymbol{p}_a - \boldsymbol{p} ||_2
$.
Then the final reward is 
$
    R = r_s+r_c+r_o+r_g+\lambda r_{f}
$,
where the $\lambda$ controls the weight of off route reward.

\begin{table}[h]
\caption{Reward Design}
\label{reward}
\begin{center}
\begin{tabular}{|c |c|}
\hline
\textbf{Reward} & \textbf{Value}\\
\hline
step penalty $r_s$ & -0.1 (move) or -0.5 (wait) \\
\hline
collision penalty $r_c$ & -5 \\
\hline
 oscillation penalty $r_o$
 &  -0.3\\
\hline
 off-route penalty $r_{f}$
 &  -$\min_{\boldsymbol{p} \in \mathcal{S}} ||\boldsymbol{p}_a - \boldsymbol{p} ||_2$\\
\hline
goal-reaching reward $r_g$
 &  30\\
\hline
\end{tabular}
\end{center}
\end{table}

\begin{algorithm} 
\caption{Multi-Agent Evolutionary Training Approach} 
\label{alg:EA} 
\begin{algorithmic}[1] 
\REQUIRE ~ Agents number $N$; discount factor $\gamma$; evolution interval $K$; evolution rate $\eta$;
\STATE Initialize agents' model weights $\Theta = \{\Theta_1,...,\Theta_N\}$
\REPEAT
\STATE Set accumulated reward $R^{(k)}_1,...,R^{(k)}_N = 0$
\STATE // \textit{update model parameters via A2C algorithm}
\FOR{$k=1,...,K$}
\FOR{each agent $i$}
\STATE Executing the current policy $\pibs_{\Theta_i}$ for $T$ timesteps, collecting action, observation and reward $\{a_i^{t}, \obs_i^t, r_i^t \}$, where $t\in [0, T]$
\STATE Compute return $R_i = \sum_{t=0}^{T} \gamma^tr_i^t$
\STATE Estimate advantage $\hat{A}_i = R - V^{\pibs_{\Theta_i}}(\obs_i)$
\STATE Compute gradients $
    \nabla_{\Theta_i} J = \mathbb{E}[\nabla_{\Theta_i} \log \pibs_{\Theta_i}\hat{A}_i]$
\STATE Update $\Theta_i$ based on gradients $\nabla_{\Theta_i} J$
\ENDFOR
\STATE $R^{(k)}_i = R^{(k)}_i + R_i$
\ENDFOR
\STATE Normalize accumulated reward to get $\Bar{R}^{(k)}_1,...,\Bar{R}^{(k)}_N$
\STATE Find maximum reward $\Bar{R}_{j}^{(k)}$ with agent index $j$
\STATE // \textit{Evolutionary selection}
\FOR{ each agent $i$}
\STATE Sample $m$ from uniform distribution between $[0,1]$
\STATE Compute evolution probability $p_i = 1-\frac{\exp (\eta\Bar{R}_i^{(k)})}{\exp (\eta\Bar{R}_j^{(k)})}$
\IF{$m < p_i$}
\STATE $\Theta_i \xleftarrow{} \Theta_j$
\ENDIF
\ENDFOR
\UNTIL converged
\end{algorithmic}
\end{algorithm}

\subsection{Model Architecture}
\label{archi}
We use deep neural networks to approximate the policy and the value function in our A2C method. 
The model architecture is illustrated in Fig.~\ref{fig:framework}. We have two input sources to be processed independently before being concatenated as a combined feature. The first one is the three channels image represented observation, which has been introduced in section ~\ref{obs}. The image channels are passed through several blocks, which contain convolution layers, and max-pooling layers. After the final block, the extracted feature will be flattened to one feature embedding. 

We notice that reinforcement learning may hardly solve long-term tasks to get the reaching goals rewards \cite{eysenbach2019search}. Therefore, instead of using final goals as one input source, we use the waypoint coordinates as sub-goals of our task, which is computed by the global planner. More specifically, the global planner, which is the A* planner in our case, will generate a reference path from the start point to the goal. Then the agent will choose waypoints on the reference path based on a certain distance interval threshold and attempt to reach them one by one. Once the agent approaches its current waypoint goal within a pre-defined range, it will begin to head to the next waypoint. 

The currently selected waypoint can be viewed as a sub-goal. It will be passed through a fully connected layer, and then be fed together with the observation feature embedding to two shared fully connected layers. The output feature of the two shared layers will then be passed through two separate neural networks. The lower one is a two-layers policy network with softmax activation, which produces the probability of choosing each action. The upper one is the value function network, which outputs the expected value of the current state.

\subsection{Multi-Agent Evolutionary Reinforcement Learning}
Although reinforcement learning has achieved great success in many single-agent tasks \cite{mnih2013playing}, it is still hard to directly apply those methods to the multi-agent case.
One challenge is the scalability issue: as the number of agents grows, the environment becomes more complicated and the variance of policy gradients may grow exponentially \cite{lowe2017multi}.

Inspired by evolutionary algorithm that has been successfully applied to many optimization problems \cite{simon2013evolutionary}, we adopt a decentralized evolutionary approach based on A2C algorithm, which can be applied to arbitrary number of agents training procedure. Evolutionary algorithm usually contains three stages: crossover, mutation and selection. Let's denote the model parameters of agent $i$ as $\Theta_i$. We firstly initialize $N$ agents with random weights for their own model. Then the mutation process begins by training each agent's model separately using A2C algorithm. After $k$ episodes training, agent $i$ will accumulate the rewards over the last $k$ episodes, and we denote it as $R_i^{(k)}$. Denote $R_{max}^{(k)} = \max_{i \in \{1,...,N\}} R_i^{(k)}$ and $R_{min}^{(k)} = \min_{i \in \{1,...,N\}} R_i^{(k)}$. We normalize the accumulated reward for agent $i$ by:
$
    \Bar{R}_i^{(k)} = \frac{R_i^{(k)}}{R_{max}^{(k)} - R_{min}^{(k)}}
$.
Assume agent $j$ has the maximum normalized reward $\Bar{R}_{j}^{(k)} = \max_{i \in \{1,...,N\}} \Bar{R}_i^{(k)}$, then we start the crossover and selection stages. Each agent $i$ has the probability $p_i$ to reserve its original model weights  and $1-p_i$ probability to replace its weights with the weights of agent $j$. The probability is calculated by
$
    p_i = 1-\frac{\exp (\eta\Bar{R}_i^{(k)})}{\exp (\eta\Bar{R}_j^{(k)})}
$.
where $\eta$ controls the evolution rate. Larger $\eta$ means agents with lower rewards are more likely to be updated. The core idea of our evolutionary method is very simple: gradually eliminate bad policies while maintaining good ones. The full MAPPER training process is shown in Algorithm~\ref{alg:EA}.

\section{Experiment and Discussion}
\label{sec:experiment}
\subsection{Experiment Settings}
\begin{figure}[htb]
\centering
\includegraphics[width=8.5cm]{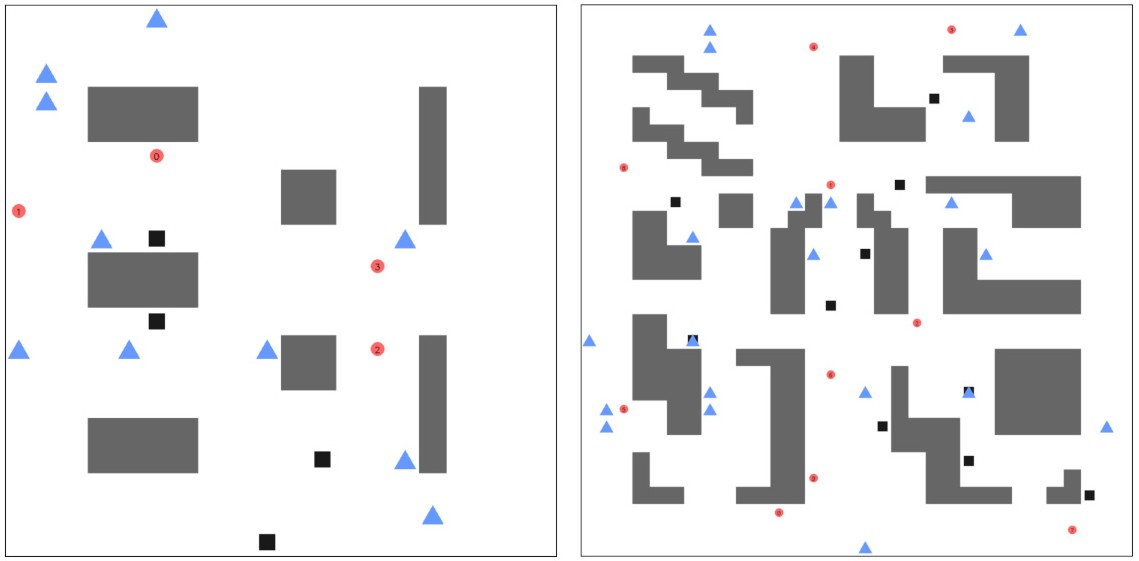}
\caption{Grid world simulation environment demonstration.}
\label{fig:map}
\end{figure}

We evaluate our approach in grid world simulation environment, just as Fig.~\ref{fig:map} shows. Gray blocks are static obstacles and black blocks are agents' goals. Orange circles represent agents. Each agent has a 7-grid sensing range in our experiment setting, which means the size of the observation image is $15 \times 15 \times 3$. Blue triangles present dynamic obstacles, where each dynamic obstacle will navigate to a randomly selected goal using LRA* algorithm~\cite{coop}. To increase the dynamic obstacle movement pattern diversity, we randomly select 50\% dynamic obstacles that will ignore the presence of surrounding agents, which would be more challenging for our agents because of their non-cooperative nature.

Existing centralized multi-agent path planning methods, such as conflict based search \cite{sharon2015conflict}, break down in mixed dynamic environments because of the unpredictable nature of non-cooperative moving obstacles. Therefore, we resort to decoupled reaction-based planning approaches. One benchmark we adopt is a modified local repair A* (LRA*) algorithm that re-plans at every time step, where we replace A* with D* lite \cite{koenig2005fast} implementation because the latter is more computationally efficient in dynamic environments \cite{coop}.
Each LRA* agent takes into account local observation, updates the cost map accordingly, and searches for a route to the destination. 
Then, a coordinator that has access to every agent's future plan resolves conflicts between agents and adjust these paths. LRA* behaves similarly to our MAPPER method in that they both react based on local observation, but note that we do not require access to all agents' future plan information.

Another baseline is PRIMAL \cite{sartoretti2019primal}, a reinforcement learning-based decentralized planner. We modify the original PRIMAL to adjust to our experiment setting because the original model and observation representation do not consider non-cooperative dynamic obstacles. More specifically, we use the same observation representation and network architecture as ours, but keep the original A3C training procedures and goal-conditioned approach as PRIMAL, so we name it PRIMAL* in the rest of the paper. 
We also conduct experiments that remove part of features of our MAPPER method, which are removing the moving dynamic obstacles' trajectory (w/o traj) and removing the global planner guidance feature (w/o guid). We evaluate the performance of each method in terms of the \textbf{success rate} in different experiment settings.

\begin{table*}[t]
\centering
\caption{Comparison of success rate over different experiment settings}
\label{compare}
\begin{tabular}{|c|c|c|c|c|c|c|c|}
\hline
\multicolumn{3}{|c|}{Environment Setting} & \multicolumn{5}{c|}{Success Rate}                              \\ \hline
map size     & agent     & dynamic obstacle    & MAPPER      & MAPPER w/o traj & MAPPER w/o guid & PRIMAL*        & LRA*           \\ \hline
20x20        & 15        & 10             & \textbf{1.0}  & 0.971    & 0.877   & 0.964   & 0.996 \\ \hline
20x20        & 35        & 30             & \textbf{1.0}  & 0.961   & 0.836 & 0.980   & 0.999 \\ \hline
20x20        & 45        & 30             & \textbf{0.999} & 0.854   & 0.607  & 0.971   & 0.997 \\ \hline
60x65        & 70        & 100            & \textbf{1.0}  & 0.256  & 0.516  & 0.352 & \textbf{1.0}   \\ \hline
60x65        & 130       & 140            & \textbf{1.0}  & 0.473   & 0.221    & 0.404 & 0.992 \\ \hline
120x130      & 150       & 40             & \textbf{0.997} &          0.324    &  0.211      & 0.389     & 0.994 \\ \hline
\end{tabular}
\end{table*}

\subsection{Training Details}
Inspired by the idea of curriculum learning \cite{forestier2017intrinsically}, we divide the whole training procedure into two stages and start from easier tasks. We begin by initializing a small population of agents and dynamic obstacles, and sample goals within a certain distance to let agents learn a short-range navigation policy. Then we increase the agents and dynamic obstacles number, and sample goals in the whole map. 

The training parameters are the same for MAPPER and its variants. We set off-route penalty weight $\lambda = 0.3$, the evolution rate and interval $\eta=2, K=50$, the discount factor $\gamma=0.99$, and the learning rate $lr= 0.0003$. For PRIMAL*, we observe that it is sensitive to the learning rate and will not converge if we set the same learning rate as MAPPER. Therefore, we set the learning rate for PRIMAL* as 0.00005 after several experimental explorations.
For the first stage, we initialize 4 agents and 10 dynamic obstacles in a $20\times 20$ map with 7 grid goal-sample range, as shown in Fig.~\ref{fig:map} left. For the second stage, we train models with 20 agents and 30 dynamic obstacles in a more complex $32\times 32$ map without goal-sample limitation, as shown in Fig.~\ref{fig:map} right.


\subsection{Results}

\begin{figure}[t]
\centering
\includegraphics[width=8cm]{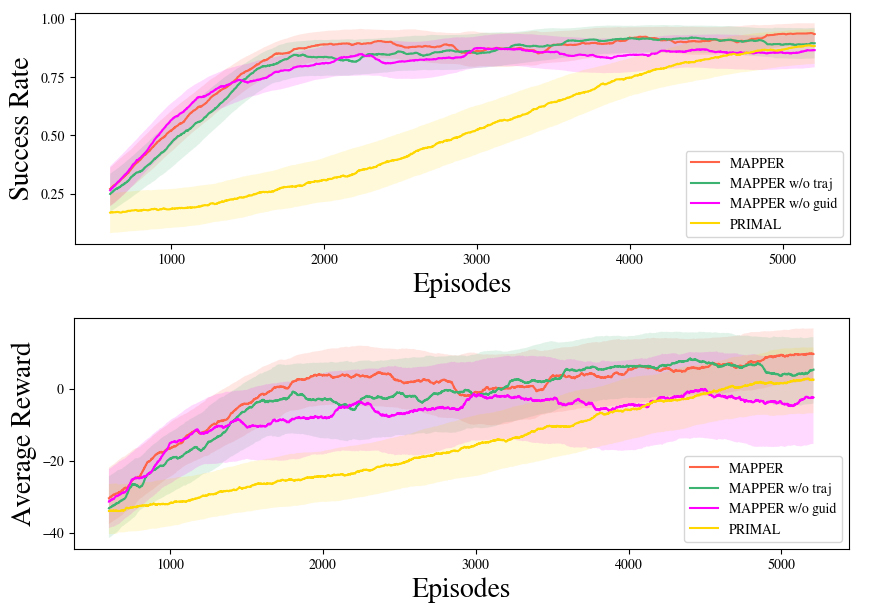}
\caption{Success rate and average reward  comparison of variants of MAPPER and PRIMAL* algorithms}
\label{fig:train}
\end{figure}

\begin{figure}[t]
\centering
\includegraphics[width=8cm]{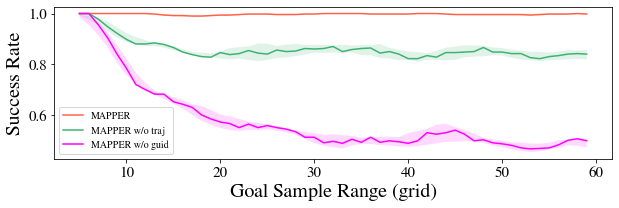}
\caption{Comparison of MAPPER and its variants with different goal sample range.}
\label{fig:goal}
\end{figure}

The training figures for the first stage are shown in Fig~\ref{fig:train}. For the second stage training, we find that MAPPER without dynamic obstacle trajectory (MAPPER w/o traj) and MAPPER without global planner guidance (MAPPER w/o guid) can hardly converge if we sample goals from the whole map, so we limit the goal sample range to 15 grids. For PRIMAL*, the proper learning rate depends on agents number because of its centralized training nature, so we keep the agents size and learning rate as in the first stage. Since the training settings are different for second stage, the training figures are not presented in Fig~\ref{fig:train}.  But from the first stage training plot, we can see MAPPER has the most stable performance (smallest variance) and fastest convergence. The final average reward and success rate of MAPPER are also superior to the other methods in comparison. 

To demonstrate the effectiveness of our observation representation, we evaluate trained models in a $65 \times 65$ size simulation environment with 10 agents, 10 dynamic obstacles, and different goal sample range. The success rate when we increase the goal range is shown in Fig.~\ref{fig:goal}.
We can see the performance of other variants of MAPPER is sub-optimal, while MAPPER agents will not be influenced by the goal range. Specifically, if we remove the global planner guidance feature, the agent's performance declines a lot when the distance to goal is increased, which means decomposing the long-range navigation task to several easier waypoint-conditioned tasks is necessary.
Though removing dynamic obstacle trajectory feature will not be influenced a lot when the goal range is changed, however, it shows worse capability to handle interactions with dynamic obstacles in a large environment.

We also evaluate the trained models as well as LRA* in various environment settings without goal sample range limitation to see their generalization capability. The performance is shown in Table~\ref{compare}.
Note that LRA* needs to access all the agents' (not dynamic obstacles) intention information and resolve conflicts before taking actions, while MAPPER only needs local observations. 
We observe that in simple tasks where only a few moving obstacles are around the MAPPER agent, the agent will behave similar to following the reference path from the global planner. However, when the dynamic obstacle density is increased and the reference path is blocked, MAPPER agent performs aggressively to get out of surrounding obstacles and then moves towards its goal. We can see the success rate for MAPPER is the highest and is consistently above 0.99 in various experiment settings.

The MAPPER variant without dynamic obstacle trajectory works well when there are few dynamic obstacles but performs poorly when the complexity of the environment increases. It can be seen that the waypoints guidance is an important aspect of the MAPPER algorithm and the variant without waypoints guidance has low success rate even in a $20\times 20$ grid world with 15 agents and 10 dynamic obstacles.

\section{CONCLUSION}
\label{sec:conclusion}
This paper proposes a decentralized partially observable multi-agent path planning with evolutionary reinforcement learning (MAPPER) method to learn an effective local planning policy in mixed dynamic environments. We model dynamic obstacle's behavior with an image-based representation and decompose the long-range navigation task into many easier waypoint-conditioned sub-tasks. Furthermore, we propose a stable evolutionary training approach that could be easily scaled to large and complex environments while maintaining good convergence property compared with centralized training methods. The experiment result shows that MAPPER outperforms traditional method LRA* and learning-based method PRIMAL* in terms of success rate among various experiment settings. However, MAPPER may still collide with other agents or dynamic obstacles in complex environments in order to reach the goal. So the future direction would be to investigate safety-critical learning-based planning methods.

\section*{ACKNOWLEDGMENT}

The authors acknowledge the support from the Manufacturing Futures Initiative at Carnegie Mellon University made possible by the Richard King Mellon Foundation.
\newpage
\bibliographystyle{IEEEtran}
\bibliography{root}
\end{document}